\documentclass[11pt,a4paper]{article}
\usepackage[hyperref]{eacl2021}
\usepackage{times}
\usepackage{latexsym}
\usepackage{graphicx}
\usepackage{multirow}
\usepackage{booktabs}
\usepackage{subfig}
\usepackage{subfloat}

\usepackage{microtype}

\aclfinalcopy %

\title{Improving Robustness by Augmenting Training Sentences with Predicate-Argument Structures}

\author{Nafise Sadat Moosavi$^{\dag}$, Marcel de Boer$^{\dag}$, Prasetya Ajie Utama$^{\dag\ddag}$, Iryna Gurevych$^{\dag}$\\
  \\
  $^{\ddag}$Ubiquitous Knowledge Processing Lab (UKP-TUDA)\\
  $^{\dag}$Research Training Group AIPHES\\
  Department of Computer Science, Technische Universit{\"a}t Darmstadt\\
  \url{https://www.ukp.tu-darmstadt.de}
  }

\date{}

\begin{document}
\maketitle
\begin{abstract}
Existing NLP datasets contain various biases, and models tend to quickly learn those biases, which in turn limits their robustness. 
Existing approaches to improve robustness against dataset biases mostly focus on changing the training objective so that models learn less from biased examples.
Besides, they mostly focus on addressing a specific bias, and while they improve the performance on adversarial evaluation sets of the targeted bias, they may bias the model in other ways, and therefore, hurt the overall robustness. 
In this paper, we propose to augment the input sentences in the training data with their corresponding predicate-argument structures, which provide a higher-level abstraction over different realizations of the same meaning and help the model to recognize important parts of sentences.
We show that without targeting a specific bias, our sentence augmentation improves the robustness of transformer models against multiple biases.
In addition, we show that models can still be vulnerable to the lexical overlap bias, even when the training data does not contain this bias, and that the sentence augmentation also improves the robustness in this scenario.
We will release our adversarial datasets to evaluate bias in such a scenario as well as our augmentation scripts at \url{https://github.com/UKPLab/data-augmentation-for-robustness}. 
\end{abstract}

\section{Introduction}
\label{sect_intro}

Due to annotation artifacts, existing datasets contain certain biases.\footnote{In this work, the term bias refers to the label bias as defined by \citet{shah-etal-2020-predictive}, i.e., the conditional distribution of the target label diverges at test time based on specific attributes of the training data.} Models often rely on these biases to perform well on the corresponding evaluation set, which also includes similar biases \citep{N18-2017,poliak-etal-2018-hypothesis,mccoy-etal-2019-right,gardner2020evaluating}.
As a result, the model learns the spurious patterns in the data instead of the intended phenomena of the dataset, which in turn limits the robustness and makes models vulnerable against adversarial evaluations \citep{mccoy-etal-2019-right,nie2019analyzing}.
The adversarial evaluation sets consist of counterexamples in which relying on the bias results in incorrect predictions.
Overcoming such biases is an important challenge in developing robust NLP models.

The majority of existing works improve the robustness against a given bias by proposing new methods or training paradigms \citep{he-etal-2019-unlearn,clark-etal-2019-dont,mahabadi2019simple,utama2020mind,utama-etal-2020-unknown,wu2020improving}.
The common component in such methods is a bias model that is trained to detect training examples that can be solved only using a bias.
This information is then used for ignoring or down-weighting biased training examples \citep{he-etal-2019-unlearn,clark-etal-2019-dont,mahabadi2019simple}, or tuning the confidence of the model on such examples \citep{utama2020mind}.
While these methods are very effective in improving the robustness against the targeted bias, they have two shortcomings: 
\begin{itemize}
	\item They mostly target a specific bias and discourage the model from learning that bias. However, they may bias the model in other unwanted directions and therefore hurt the overall robustness. \footnote{The concurrent work of \citet{utama-etal-2020-unknown} and \citet{wu2020improving} are the exceptions to this trend, in which they address multiple biases together and show that their debiasing methods improve the overall robustness.}  
	\item They are only applicable to scenarios in which the training examples contain the bias. However, as we show in this paper, a model can still be vulnerable to a specific bias even if the training examples do not explicitly exhibit that bias.
\end{itemize} 

An alternative approach is to augment the training data with additional counterexamples for the bias \citep{mccoy-etal-2019-right,elkahky-etal-2018-challenge}. This may also result in overfitting to the augmented counterexamples and hurting the overall robustness \citep{nie2019analyzing}.

In this paper, we propose to augment existing training sentences with their corresponding predicate-argument structures.
The motivation of using predicate-argument structures
is to provide a higher-level abstraction over
different surface realizations of the same underlying
meaning. 
We examine the impact of this linguistic augmentation on pre-trained transformers, e.g., BERT \cite{bert}, which achieve state-of-the-art performances on numerous NLP datasets.
The addition of predicate-argument structures to input sentences helps the model
to recognize and focus on more important parts of sentences, and therefore, to learn a different attention pattern.
The findings of this paper are as follows:
\begin{itemize}
 \item We show that the model may still be vulnerable to a specific bias even when training examples of the target task do not contain that bias. We propose new adversarial sets for evaluating the robustness of models in such scenarios based on the SWAG dataset and the lexical overlap bias.  Lexical overlap is a common bias in various NLP datasets, e.g., natural language inference
  \citep{mccoy-etal-2019-right}, or question answering \citep{Jia_2017}.
  We show that the performance of pre-trained transformers that are fine-tuned on the SWAG dataset \citep{zellers-etal-2018-swag} drop below a random baseline on evaluation sets that contain this bias.
 
\item Our results show that without targeting a specific bias or adding additional training examples, the proposed sentence augmentation improves the robustness of the model against various types of biases. Besides, we show that the sentence augmentation is effective in both scenarios when the training data contains or does not contain the bias. Our approach only requires the augmentation for the training sentences and does not require any changes to the test data, and therefore, does not add any additional cost at the test time.

\item Our results emphasize the importance of evaluating the impact of debiasing methods on more than one adversarial set and with more than one base model. 

\end{itemize}

\section{Related Work}
\paragraph{Debiasing Methods.}
Existing debiasing solutions fall into one of these two categories: (1) extending the training data with additional counterexamples, and (2) proposing a new approach that recognizes biased examples in the training data and then using this knowledge during training \citep{utama2020mind,he-etal-2019-unlearn,clark-etal-2019-dont,mahabadi2019simple,utama-etal-2020-unknown,wu2020improving}.

The first type of solution is to identify the bias and augment the training data with counterexamples in which relying on the targeted bias results in incorrect predictions.
While augmenting the training data with counterexamples improves the results on the targeted bias, it may hurt the overall robustness \citep{nie2019analyzing}.
As mentioned by \citet{jha2020does}, while augmentation with counterexamples helps the model to unlearn the targeted bias, it is unlikely that it encourages the model to rely on more generalizable features of the data.

The approaches of the second category first use a bias detection model for recognizing training examples that contain the bias. They then either (a) train an ensemble of the bias model and a base model so that the base model only learns from non-biased examples \citep{he-etal-2019-unlearn,clark-etal-2019-dont,mahabadi2019simple}, (b) change the importance of the biased training examples in the training objective \citep{schuster-etal-2019-towards,mahabadi2019simple}, or (c) change the confidence of the model on biased examples \citep{utama2020mind}.

The shortcoming of existing debiasing methods is that they mostly model a single bias and only evaluate the impact of the proposed method on the adversarial evaluation set of the targeted bias.
Therefore, while they improve the performance on the targeted adversarial sets, they may hurt the overall robustness.
The recent work of \citet{utama-etal-2020-unknown} and \citet{wu2020improving} are the exceptions in which they show that their proposed debiasing frameworks improve the overall robustness, and hence the generalization across different datasets in natural language understanding and question answering, respectively.
\citet{utama-etal-2020-unknown} propose a new framework that automatically recognizes biased training examples and does not require predefining bias types. Therefore, the recognized biased examples may contain various bias types. 
\citet{wu2020improving} propose a framework for modeling multiple known biases concurrently. To do so, they propose to combine two bias weights in the training objective including (a) a dataset-level weight indicating the strength of the bias in the datasets, and (b) an example-level weight indicating the strength of the bias in a training example.
The common finding in both works is that debiasing based on multiple biases is a key factor in improving the overall robustness.

Compared to existing debiasing methods:
\begin{itemize}
	\item Our proposed approach does not include an additional model to recognize biased examples or additional training examples.
As we show in Section~\ref{robustness_results}, since it does not target any specific bias, it improves the robustness of the baseline model against multiple biases.
\item Since it does not require recognizing biased examples, it is applicable to improve robustness against biases that do not exist in the training examples. 
\end{itemize}

\paragraph{Using Linguistic Structures for Neural Models.}
The use of linguistic information in recent neural models is not very common. 
The use of such information has been mainly investigated for tasks in which there is a clear relation between the linguistic features and the target task.
For instance, various neural models use syntactic information for the task of semantic role labeling (SRL) \cite{roth-lapata-2016-neural,marcheggiani-titov-2017-encoding,strubell-etal-2018-linguistically,swayamdipta-etal-2018-syntactic}, which is closely related to syntactic relations, i.e., some arcs in the syntactic dependency tree can be mirrored in semantic dependency relations.

\newcite{marcheggiani-titov-2017-encoding} build a graph representation from the input text using their corresponding dependency relations and use graph convolutional networks (GCNs) to process the resulting graph for SRL.
They show that the incorporation of syntactic relations improves the in-domain but decreases the out-of-domain performance.

Similarly, \newcite{cao-etal-2019-qa} and \newcite{dhingra-etal-2018-neural} incorporate linguistic information, i.e., coreference relations, in their model and show improvements in in-domain evaluations.

\newcite{strubell-etal-2018-linguistically} use linguistic information, i.e., dependency parse, part-of-speech tags, and predicates for SRL using a transformer-based encoder \cite{NIPS2017_7181}.
They make use of this linguistic information by (1) using multi-task learning, and (2) supervising the neural attention of the transformer model to predict syntactic dependencies.
They use gold syntax information during training and predicted information during the test time.
Their model substantially improves both in-domain and out-of-domain performance in SRL.
However, these results are then outperformed by a simple BERT model without using any additional linguistic information \cite{shi2019simple}.

\newcite{moosavi-strube-2018-using} examine the use of various linguistic features, e.g., syntactic dependency relations and gender and number information, as additional input features to a neural coreference resolver. They show that using informative linguistic features substantially improves the generalization of the examined model.

In a similar direction, \newcite{moosavi2019improving} improve the robustness by enhancing the input representations. They did so by adding a set of simple features to the input where the input is a pair of text sequences and show that it improves generalization across similar datasets and tasks. 

All the above approaches require additional linguistic information, e.g., syntax, both during the training and the test time.
\newcite{swayamdipta-etal-2018-syntactic}, on the other hand, only make use of the additional syntactic information during training. They use multi-task learning by considering syntax parsing as an auxiliary task and minimizing the combination of the losses of the main and auxiliary tasks. 
They use syntactic information for the tasks of SRL and coreference resolution.
They show that this information slightly improves the in-domain performance.
In this work, we do not change the loss function and only augment the input sentences of the training data.
The advantage of our solution is that it does not require any changes in the model or its training objective. It can be applied to all the transformer-based models without changing the training procedure.

\paragraph{Using Predicate-Argument Structures.}
Predicate-argument structures have been used for improving the performance of downstream tasks like machine translation \cite{liu-gildea-2010-semantic,bazrafshan-gildea-2013-semantic}, reading comprehension \cite{berant-etal-2014-modeling,wang-etal-2015-machine}, and dialogue systems \cite{turdialogue,chen-et-al:2013:PAPERS}.
However, these approaches are based on pre-neural models.

The proposed model by \newcite{marcheggiani-etal-2018-exploiting} for neural machine translation is a sample neural model that incorporates predicate-argument structures.
Unlike this work, \newcite{marcheggiani-etal-2018-exploiting} incorporate these linguistic structures at the model-level. They add two layers of semantic GCNs on top of a standard encoder, e.g., convolutional neural network or bidirectional LSTM. The semantic structures are used for determining nodes and edges in the GCNs.
In this work, however, we incorporate these structures at the input level, and only for the training data. Therefore, we can use the state-of-the-art models without any changes.

Overall, this work differs from the related work because (1) it evaluates the use of predicate-argument structures for improving the robustness of transformer-based models on natural language understanding tasks, and (2) it uses these structures at the input level to extend raw inputs, (3) it only employs this information during training, and (4) it requires no changes in the model or the training procedure.

\section{Augmenting Input Sentences with Predicate-argument Structures}
We augment the raw text of each input sentence in the training data with its corresponding predicate-argument structures.
We use the PropBank-style semantic role labeling model of \newcite{shi2019simple}, which has state-of-the-art results on the CoNLL-2009 dataset.
We specify the beginning of the augmentation by the \texttt{[PRD]} special token that indicates that the next tokens are the detected predicate.\footnote{Special tokens are atomic, i.e., they are not split by the tokenizer.}
We then specify the ARG0 and ARG1 arguments, if any, with \texttt{[AG0]} and \texttt{[AG1]} special tokens, respectively.
The end of the detected predicate-argument structure is also specified by the \texttt{[PRE]} special token.
If more than one predicate is detected for a sentence, we at most add the first three detected predicate-argument structures.\footnote{In our preliminary experiments, we found out that this setting works better than adding all of them.}
Figure~\ref{fig:sentence_pred_arg_example} shows an example for an augmented sentence.

\begin{figure}[htb]
    \centering
\fbox{\begin{minipage}{0.998\columnwidth}

\textbf{Original:} Someone takes the drink, then holds it.  

\textbf{Augmented:} Someone takes the drink, then holds it.  \texttt{[PRD]} takes \texttt{[AG0]} Someone \texttt{[AG1]} the drink \texttt{[PRE]}   \texttt{[PRD]} holds \texttt{[AG0]} Someone \texttt{[AG1]} it \texttt{[PRE]}
\end{minipage}}    
\caption{Augmenting the text of an input sentence with its predicate-argument structures.}
    \label{fig:sentence_pred_arg_example}
\end{figure}

\section{Impact of Sentence Augmentation on Improving Robustness }
In this section, we explain the datasets and models that we use in our experiments, as well as the result of the sentence augmentation compared to other recent debiasing methods.

\subsection{Training Data}
As mentioned, we have evaluated the impact of the proposed augmentation on two different settings: (1) when the training data contains the investigated biases, and (2) when training examples do not explicitly contain the bias.

\paragraph{Training Data with Biases (MultiNLI).} 
For evaluating the impact of data augmentation when the training examples are biased, we use MultiNLI \cite{N18-1101} for training, which is the largest available dataset for Natural Language Inference (NLI).
Given a premise and a hypothesis, NLI is the task of  determining whether the hypothesis is entailed, contradicts, or is neutral to the premise.

Various studies show that similar to many other NLP datasets, MultiNLI contains various biases.
For instance, hypothesis sentences may contain words that are highly associated with a target label, regardless of the premise \citep{N18-2017,S18-2023}.
This bias is referred to as the hypothesis-only bias.
Another well-known bias in MultiNLI is the lexical overlap bias, i.e., the label of most premise-hypothesis pairs with overlapping words is entailment.

\paragraph{Training Data without the Bias (SWAG).}
For the second setting, we evaluate the impact of the augmentation to improve the robustness against the lexical overlap bias. Lexical overlap is a common bias in various NLP datasets, e.g., NLI
\citep{mccoy-etal-2019-right}, duplicate question detection \citep{zhang-etal-2019-paws}, or question answering \citep{jia-liang-2017-adversarial}.%
We use the SWAG dataset \citep{zellers-etal-2018-swag} as the training data.

Given a premise about a situation, the task of the SWAG dataset, i.e., grounded commonsense reasoning, is to reason about what is happening and to predict what might come next.  
The task is modeled as a multiple choice answer selection.
For instance, ``The tutorial starts by showing each part of the drum set up close'' is a correct ending for the premise ``A man in a black polo shirt is sitting in front of an electronic drum set''.

If we train the bias model of \citet{clark-etal-2019-dont} for solving the task only based on lexical overlap features, it only achieves 26\% accuracy on SWAG, which is around the random baseline, while it
achieves 65\% accuracy on MultiNLI.\footnote{The details of this bias model is reported in the supplementary materials.}
This indicates that the examples in the SWAG dataset are not affected by this bias.
Please note that the SWAG dataset may also contain various biases. However, it does not contain the lexical overlap bias.

\subsection{Evaluation Sets}
In this section, we discuss the adversarial evaluation sets that we use to evaluate the robustness of the model.
Apart from the adversarial sets, which are the \textbf{out-of-distribution} evaluation sets compared to the training data distribution, we also report the performance on the corresponding development set as the \textbf{in-domain} performance. 
\subsubsection{MultiNLI Evaluation Sets} 
We use the following adversarial sets for evaluating models that are trained on MultiNLI:

\paragraph{MultiNLI \textbf{Hard}:} \citet{N18-2017} introduce a hard split for MultiNLI evaluation sets in which models cannot predict the correct labels using the hypothesis-only bias.

\paragraph{HANS:} \citet{mccoy-etal-2019-right} create this dataset for evaluating the lexical overlap bias. Sentence pairs in HANS include various forms of lexical overlap, namely \emph{lexical overlap}, \emph{subsequence}, and \emph{constituent}.

In the \emph{lexical overlap} subset, all words of the hypothesis appear in the premise.
For instance, ``The doctor was paid by the actor'' and ``The doctor paid the actor'' sentence pair belong to this subset.

The \emph{subsequence} subset contains hypotheses, which are a contiguous subsequence of their corresponding premise.
``The doctor near the actor danced'' and ``The actor danced'' are sample sentence pairs from this subset.

Finally, in the \emph{constituent} subset, hypotheses are a complete subtree of the premise.
For example, ``If the artist slept, the actor ran'' and ``The artist slept'' belong to this subset.

\paragraph{Stress Test:} \citet{naik-etal-2018-stress} provide adversarial evaluation sets based on weaknesses of state-of-the-art NLI models. We use \emph{negation}, \emph{word overlap}, and \emph{length mismatch} sets from the stress test, in which a tautology is added at the end of the premise or hypothesis in MultiNLI.

\begin{itemize}
\item \textbf{Negation}: the tautology ``and false is not true'' is added to the end of all the hypothesis sentences in the MultiNLI development set for creating this evaluation set. The presence of the negation word ``not'' may confuse the model to predict contradiction.
\item \textbf{Word Overlap}: For creating this evaluation set, \citet{naik-etal-2018-stress} append the tautology ``and true is true'' to the end of all the hypothesis
sentences in the MultiNLI development set.
\item \textbf{Length Mismatch}: the tautology ``and true is true'' is appended five times to the
end of the premise sentences in the MultiNLI development set for creating this adversarial evaluation set.
\end{itemize}

\subsubsection{SWAG Evaluation Sets} 
We created three different adversarial datasets based on the SWAG development set for evaluating the lexical overlap bias.
These datasets evaluate the model's understanding of (1) syntactic variations, (2) antonym relations, and (3) named entities in the presence of high lexical overlap.

The common property in all three evaluation sets is a high lexical overlap between the sentence pairs.
In all these evaluation sets, one of the incorrect endings is replaced with a new incorrect ending that has a high lexical overlap with the premise.
Since the new incorrect endings are created automatically, they may contain sentences that are not meaningful.
For instance, the syntactic variations subset contains the incorrect endings ``a key holds up someone'' and ``The last page flips to the writer'' for the  ``Someone holds up a key'' and ``The writer flips to the last page'' premises, respectively.
Humans can recognize such sentences are not meaningful, and therefore, they are not a plausible ending for given premises.
However, as we will see, because of the lexical overlap bias, the model mostly selects the new incorrect endings. 

\paragraph{Syntactic Variations:} 
\label{sect:compositional}
In this evaluation set, we take premises that contain subject-verb-object structures from the \emph{SWAG} development set.
We then construct a new negative ending by swapping the subject and object of the premise and replace one of the existing negative endings with the new one.\footnote{We use Stanford parser \citep{chen-manning-2014-fast} for detecting subjects and objects.}
This dataset includes 20K samples. It is similar to a subset of the \emph{lexical overlap} subset in HANS, as well as the adversarial evaluation that is explored by \newcite{bansal:AAAI:2019} for NLI.

\paragraph{Antonym Relations:}
\label{sect_antonym}

In this test set, we create a new negative ending by replacing the first verb of the premise with its antonym.
We use WordNet for antonym relations.
This adversarial setting is also common in NLI, e.g., \cite{naik-etal-2018-stress,glockner-etal-2018-breaking}.
As an example, ``A lot of people are \emph{standing} on terraces in a big field and people is walking in the entrance of a big stadium'' is an incorrect ending for the ``A lot of people are \emph{sitting} on terraces in a big field and people is walking in the entrance of a big stadium'' premise in this evaluation set.
This set contains 7476 samples.

\paragraph{Named Entities:}
In this adversarial dataset, a new incorrect ending is created by replacing one of the named entities of the premise with an unrelated named entity.\footnote{We use the Stanford named entity recognizer \cite{finkel-etal-2005-incorporating} for determining the named entities.}
For instance, based on the ``The reflection he sees is \emph{Harrison Ford} as someone Solo winking back at him'' premise, we create ``The reflection he sees is \emph{Eve} as someone Solo winking back at him.'' as the new incorrect ending.
This test set contains 190 samples. 

\subsection{Base Model}
We use the Bert-base-uncased model \citep{devlin-etal-2019-bert} as the base model.\footnote{We use Huggingface Transformers \cite{Wolf2019HuggingFacesTS}.}
\emph{Bert-orig} refers to the results when the model is trained on the original training data.
\emph{Bert-aug} represents the results when the base model is trained on the augmented data.
The set of all parameters are the same in \emph{Bert-orig} and \emph{Bert-aug}.
Besides, the evaluation data is the same for both \emph{Bert-orig} and \emph{Bert-aug} experiments, and their only difference is their training data.

We compare our results with the confidence regularization approach of \citet{utama2020mind} and the product-of-expert approach \cite{he-etal-2019-unlearn,clark-etal-2019-dont}.
They both use Bert-base-uncased as the base model.
\emph{CR(lex)} and \emph{CR(hypo)} refer to the confidence-regularization method when it is debiased based on the lexical overlap and hypothesis-only biases, respectively.
Similarly,  \emph{POE(lex)} and \emph{POE(hypo)} show the product-of-expert results based on the lexical overlap and hypothesis-only biases, respectively.

We use the same set of hyper-parameters\footnote{I.e., batch size=16, learning rate=2e-5, and the same set of random seeds.} 
for all the models and report the average performance using five different random seeds for all the results, i.e., Table~\ref{tab:sota_comparison}-Table~\ref{tab:swag_all}.
As reported by \citet{mccoy2019} and \citet{zhou2020curse}, the performance on the adversarial evaluation sets can highly vary given different values of hyper-parameters.\footnote{E.g., The accuracy on the non-entailment examples in the lexical overlap subset of HANS can vary between 6\% to 54\% for the BERT-base model using different random seeds.} 
Therefore, to ensure a fair comparison, results on adversarial sets should be reported using the same parameters in out-of-distribution evaluations.

\begin{table*}[!htb]
\footnotesize
	\begin{center}
\resizebox{\textwidth}{!}{%

		\begin{tabular}{ l r | rr  r|r |rrr }
				\toprule
		 & \multicolumn{1}{c}{\textbf{In-domain}} & \multicolumn{3}{c}{\textbf{HANS}} & \multicolumn{1}{c}{\textbf{Hard}} & \multicolumn{3}{c}{\textbf{Stress Test}}  \\
 \textbf{Model} & \multicolumn{1}{c}{} & \multicolumn{1}{c}{lex.} & \multicolumn{1}{c}{subs.} & \multicolumn{1}{c}{const.} & \multicolumn{1}{c}{}  & \multicolumn{1}{c}{{Negation}} & \multicolumn{1}{c}{{Overlap}} & \multicolumn{1}{c}{{Length}} \\
        \midrule
        BERT-orig & 84.2$\pm$0.3 & 62.9$\pm$7.8 & 52.1$\pm$0.9 & 56.0$\pm$1.4 & 75.3$\pm$0.6 & 55.5$\pm$0.6 & 59.5$\pm$1.1 & 81.3$\pm$0.3\\
        \midrule
        CR(lex) & \textcolor{gray}{83.7}$\pm$0.1 & 62.1$\pm$2.7 & {60.9}$\pm$3.7 & {64.2}$\pm$1.6 & \textcolor{gray}{74.8}$\pm$0.3  & 55.6$\pm$0.4 & 59.5$\pm$1.0 & 81.3$\pm$0.3 \\ 
        CR(hypo) & \textbf{84.4}$\pm$0.2 & 68.5$\pm$8.3 & {53.9}$\pm$1.2 & 56.2$\pm$0.8 & {77.2}$\pm$0.4  & \textcolor{gray}{55.1}$\pm$0.4 & 59.5$\pm$1.3 & \textbf{81.5}$\pm$0.3  \\
		
        PoE(lex) & \textcolor{gray}{82.6}$\pm$0.2 & 69.1$\pm$5.9 & \textbf{68.1}$\pm$9.6 & \textbf{68.8}$\pm$3.6 & \textcolor{gray}{73.2}$\pm$0.2& 55.6$\pm$0.4 & \textcolor{gray}{59.1}$\pm$0.8 & \textcolor{gray}{80.7}$\pm$0.4 \\
		
        PoE(hypo) & \textcolor{gray}{83.3}$\pm$0.4 & 66.6$\pm$7.3 & 53.1$\pm$1.2 & 58.8$\pm$2.3 & \textbf{77.8}$\pm$0.7 & \textcolor{gray}{55.2}$\pm$0.5 & 59.7$\pm$0.8 & \textcolor{gray}{80.5}$\pm$0.3 \\
        
                BERT-aug & \textbf{84.4}$\pm$0.1 & \textbf{69.8}$\pm$5.6 & {53.5}$\pm$1.5 & 58.3$\pm$2.0 & {75.9}$\pm$0.3  & \textbf{56.3}$\pm$0.2 & \textbf{60.2}$\pm$1.4 & \textbf{81.5}$\pm$0.3\\

        \bottomrule
		\end{tabular}
}
	\end{center}
	\caption{Comparing the impact of the augmentation to the confidence regularization (CR) \citep{utama2020mind}, and product-of-expert (POE) \citep{he-etal-2019-unlearn,clark-etal-2019-dont} methods debiased for the lexical overlap (lex) and hypothesis-only (hypo) biases. E.g., CR(lex) is the confidence regularization method debiased based on lexical overlap. All models are trained on MultiNLI with the same hyperparameters.
	Highest scored on each dataset are boldfaced. Scores that are lower that BERT-orig are marked in gray.}
	\label{tab:sota_comparison}
\end{table*}

\begin{table}[!htb]
\footnotesize
\resizebox{\columnwidth}{!}{%
	\centering
		\begin{tabular}{ l r | rrr}
				\hline
		\textbf{Model} & \multicolumn{1}{c}{\textbf{Dev.}} & \multicolumn{1}{c}{\textbf{Syntax}} & \multicolumn{1}{c}{\textbf{Antonym}} & \multicolumn{1}{c}{\textbf{NEs}} \\
        \hline
        BERT-base & \textbf{81.1}$\pm$0.1 &27.7$\pm$0.9 & 18.3$\pm$1.7 & 7.9$\pm$1.1\\
        BERT-aug & 79.1$\pm$0.3 & \textbf{47.1}$\pm$1.4 & \textbf{36.3}$\pm$1.6 & \textbf{15.9}$\pm$1.7\\ %
        \hline
		\end{tabular}
	}
	\caption{Results on SWAG and it's adversarial sets. }
	\label{tab:swag_bert}
\end{table}

\subsection{Results} 
\label{robustness_results}
Table~\ref{tab:sota_comparison} and Table~\ref{tab:swag_bert} show the results of the examined models on all evaluation sets of MultiNLI and SWAG datasets, respectively.
MultiNLI and its corresponding adversarial evaluation sets, i.e., HARD and Stress Test, contain two subsets, matched and mismatched. 
Sentence pairs in the matched subsets are from the same domain as those of the training data while they are from different domains in the mismatched subsets.
We have reported the results on the matched subsets in Table~\ref{tab:sota_comparison}. The results on the mismatched subsets are included in the supplementary material, and they follow the same pattern.

The confidence regularization and product-of-expert debiasing methods model a single bias at a time and train a bias detection model to detect training examples that can be solved by only using biased features.
Therefore, they are only applicable when training examples contain the examined bias, and they are not used for the experiments of Table~\ref{tab:swag_bert}.

Based on the results of Table~\ref{tab:sota_comparison}:
\begin{itemize}
	\item The model that is debiased for a specific bias has a higher accuracy on the corresponding adversarial evaluation set, i.e., \emph{POE(lex)} has the highest average score on HANS. However, they can reduce the performance on the non-targeted evaluation sets, e.g., \emph{POE(lex)} results are below the baseline on HARD and Stress Test evaluation sets.
	\item The use of sentence augmentation results in consistent improvements across the examined evaluation sets.
\end{itemize}
Based on the results of Table~\ref{tab:swag_bert}, we see that while BERT has very high accuracy on the SWAG development set, its performance drops below a random baseline on the lexical overlap adversarial sets.
This indicates that the model is biased towards selecting the endings which have a high lexical overlap to the premise, while the training data does not contain this bias.
Besides, we see that augmenting the training data improves the accuracy on the adversarial sets from 8-20 points.

\begin{figure*}[!htb]
   \centering
   
   \subfloat{\includegraphics[width=.3\textwidth]{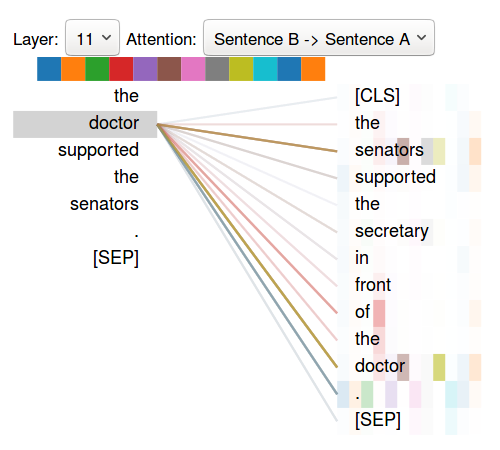}}\quad
   \subfloat{\includegraphics[width=.3\textwidth]{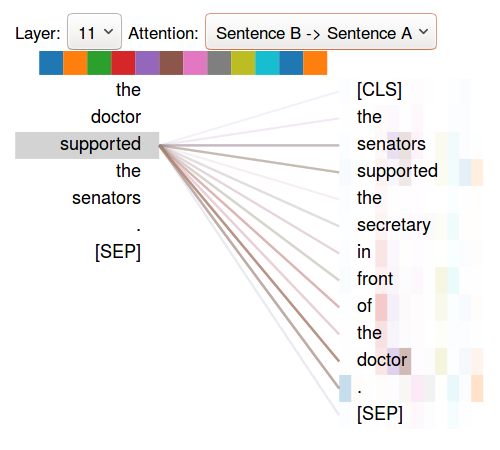}}
   \subfloat{\includegraphics[width=.3\textwidth]{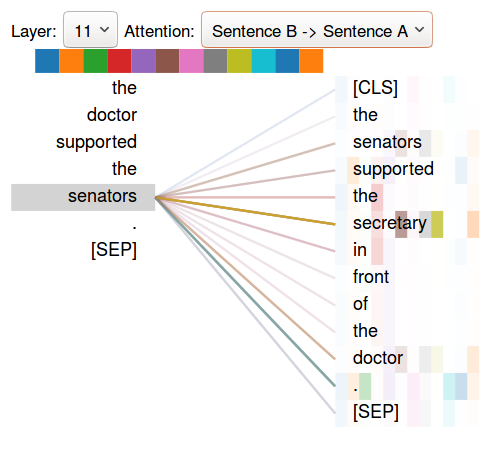}}
   \\
   \subfloat{\includegraphics[width=.3\textwidth]{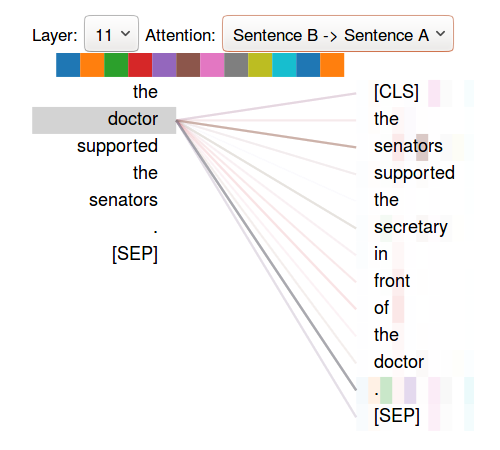}}\quad
   \subfloat{\includegraphics[width=.3\textwidth]{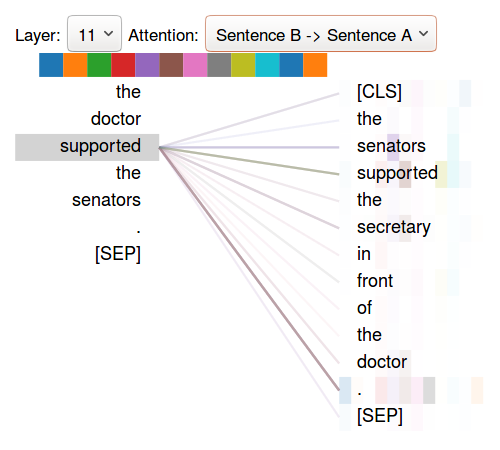}}
   \subfloat{\includegraphics[width=.3\textwidth]{figures/hans_prop_orig_arg2.png}}
   \caption{BERT attention weights on an example from the HANS dataset based on \emph{original} (top weights) and \emph{augmented} (bottom weights) training. Attention weights are visualized using BertViz \cite{vig2019transformervis}. They highlight the attention between the hypothesis and premise words and for the predicate-argument structures of the hypothesis.}
   \label{fig:ln_preposition}
\end{figure*}

\paragraph{An example of the attention pattern with and without sentence augmentation}
Figure~\ref{fig:ln_preposition} shows the difference of the BERT attention weights, using BertViz\footnote{\url{https://github.com/jessevig/bertviz}} \cite{vig2019transformervis}, on an example from the HANS dataset.
In this example, the premise and hypothesis are ``The senators supported the secretary in front of the doctor.'' and	``The doctor supported the senators.'', respectively.

For instance, for the predicate ``supported'' in the hypothesis, the BERT model that is trained on \emph{augmented} data  (bottom subfigure), has high attention weights on ``senators'', ``supported'', and ``secretary'', while for \emph{original} the attention weights of this predicate are more distributed.
Similarly, for the subject ``doctor'' in the hypothesis, \emph{augmented} mainly attends to the corresponding subject in the premise, i.e., ``senators''.

\begin{table*}[!htb]
\footnotesize
	\begin{center}
\resizebox{\textwidth}{!}{%

		\begin{tabular}{ l r | rr  r|r |rrr }
				\hline
		 & \multicolumn{1}{c}{\textbf{In-domain}} & \multicolumn{3}{c}{\textbf{HANS}} & \multicolumn{1}{c}{\textbf{Hard}} & \multicolumn{3}{c}{\textbf{Stress Test}}  \\
 \textbf{Model} & \multicolumn{1}{c}{} & \multicolumn{1}{c}{lex.} & \multicolumn{1}{c}{subs.} & \multicolumn{1}{c}{const.} & \multicolumn{1}{c}{}  & \multicolumn{1}{c}{{Negation}} & \multicolumn{1}{c}{{Overlap}} & \multicolumn{1}{c}{{Length}} \\
        \hline
        XLNET-orig & 86.6$\pm$0.1 & 70.2$\pm$1.3 & 53.6$\pm$0.9 & 66.5$\pm$2.2 & 79.4$\pm$0.8 & 57.6$\pm$1.2 & 62.7$\pm$0.7 & 84.0$\pm$0.4 \\
        XLNET-aug & 86.1$\pm$0.3 & \textbf{71.5}$\pm$1.1 & \textbf{55.4}$\pm$1.2 & \textbf{66.6}$\pm$4.7 & 79.0$\pm$0.6 & \textbf{60.6}$\pm$1.5 & \textbf{67.5}$\pm$4.1 & \textbf{84.1}$\pm$0.1 \\
        \midrule
        RoBERTa-orig & 87.5$\pm$0.1 &  83.3$\pm$2.3 & 64.6$\pm$1.0 & 68.1$\pm$3.3 & 80.6$\pm$0.1 & 57.3$\pm$0.4 & 65.5$\pm$1.8 & 85.0$\pm$0.2 \\
        RoBERTa-aug & 87.3$\pm$0.2 &  82.1$\pm$1.7 & 61.8$\pm$0.8  & 66.3$\pm$3.5 & 80.3$\pm$0.2  & 57.2$\pm$0.5 & 63.7$\pm$1.2 & 85.0$\pm$0.1 \\
        \hline
		\end{tabular}
}
	\end{center}
	\caption{Impact of linguistic augmentation on the XLNET and RoBERTa models trained on MultiNLI. 
	Accuracy scores that are higher than the baseline are boldfaced.}
	\label{tab:xlnet-roberta}
\end{table*}

\section{Are the improvements model-agnostic?}
We also evaluate the impact of our augmentation on other transformer models including {XLNET} \citep{yang2019xlnet} and RoBERTa \citep{liu2019roberta}.

The differences of the examined transformer models are as follows: {BERT} is jointly trained on a masked language modeling task and a next sentence prediction task. 
It is pre-trained on the BookCorpus and English Wikipedia.
{XLNET} \cite{yang2019xlnet} is trained with a permutation-based language modeling objective for capturing bidirectional contexts. The XLNet-base model is trained with the same data as BERT-base.
The RoBERTa model \cite{liu2019roberta} has the same architecture as BERT. However, it is trained with dynamic masking and without the next sentence prediction task. It is also trained using larger batch-size, vocabulary size, and training data. 

Table~\ref{tab:xlnet-roberta} and Table~\ref{tab:swag_all} present the results when the models are trained on the original vs.\ augmented MultiNLI and SWAG training data, respectively.
\begin{table}[htb]
\footnotesize
\resizebox{\columnwidth}{!}{%
	\centering
		\begin{tabular}{ l r | rrr }
				\hline
		\textbf{Model} & \multicolumn{1}{c}{\textbf{Dev.}} & \multicolumn{1}{c}{\textbf{Syntax}} & \multicolumn{1}{c}{\textbf{Antonym}} & \multicolumn{1}{c}{\textbf{NEs}}\\
        \hline
        XLNET-orig & {80.2}$\pm$0.2 & 37.0$\pm$1.6 & 25.4$\pm$1.7 & 17.1$\pm$2.8 \\
        XLNET-aug & 77.8$\pm$0.3 & \textbf{63.3}$\pm$1.1 & \textbf{50.5}$\pm$1.2 & \textbf{36.7}$\pm$3.8\\ \midrule
        RoBERTa-orig & {83.7}$\pm$0.2 & 49.1$\pm$2.4 & 31.7$\pm$2.0 & 24.1$\pm$1.3 \\
        RoBERTa-aug & 82.0$\pm$0.3 & \textbf{60.9}$\pm$1.1 & \textbf{47.7}$\pm$1.3 & \textbf{38.6}$\pm$4.0\\
        \hline
		\end{tabular}
	}
	\caption{Impact of sentence augmentation on XLNET and RoBERTa trained on SWAG. 
	}
	\label{tab:swag_all}
\end{table}

The results show that 
when the model itself is relatively robust and has high performance on various evaluation sets, which is the case for RoBERTa trained on MultiNLI, augmenting the training data will not have a positive impact.
While several recent debiasing methods are model-agnostic, e.g., \citep{utama2020mind,mahabadi2019simple,clark-etal-2019-dont}, they are evaluated using only the BERT base model.
This result indicates the importance of evaluating model-agnostic methods on more than one model.

\section{Conclusions}
We propose a new approach for improving the robustness of transformer models by augmenting the training sentences with their corresponding predicate-argument structures.
We show that without targeting any specific bias, sentence augmentation improves the robustness against different types of biases.
Sentence augmentation is independent of the underlying task and model and therefore applies to different tasks and settings.
The augmentation only applies to the training examples, and therefore, it does not add any additional complexity at the test time.
We evaluate the impact of the proposed augmentation on the natural language inference and grounded common sense reasoning tasks. However, this work opens new research directions on improving robustness by using better linguistically-informed input representations, rather than simply using raw texts. 
To ensure improved robustness, we encourage the community to evaluate their debiasing methods on (1) more than one evaluation set, (2) in a wider setting in which the bias does or does not exist in the training data, and (3) with more than one base model. 
\section*{Acknowledgments}
We would like to thank Kevin Stowe, Jan-Christoph Klie, Zahra Ahmadi, and the anonymous reviewers for their constructive feedbacks. This work is funded by the German Research Foundation through the research training group AIPHES (GRK 1994/1) and by the German Federal Ministry of Education and Research and the Hessian Ministry of Higher Education, Research, Science and the Arts within their joint support of the National Research Center for Applied Cybersecurity ATHENE.
\bibliography{paper}
\bibliographystyle{acl_natbib}

\appendix
\begin{table*}[htb]
\footnotesize
	\begin{center}

		\begin{tabular}{ l r |r |rrr }
				\toprule
		 & \multicolumn{1}{c}{\textbf{In-domain}}  & \multicolumn{1}{c}{\textbf{Hard}} & \multicolumn{3}{c}{\textbf{Stress Test}} \\
 \textbf{Model} & \multicolumn{1}{c}{} & \multicolumn{1}{c}{} & \multicolumn{1}{c}{{Negation}} & \multicolumn{1}{c}{{Overlap}} & \multicolumn{1}{c}{{Length}} \\
        \midrule
        BERT-base & {84.7}$\pm$0.2 & 77.1$\pm$0.2 & 56.1$\pm$0.7 & 59.1$\pm$0.6 & 82.3$\pm$0.3\\
        \midrule
        CR (lex) & \textcolor{gray}{84.2}$\pm$0.2 & \textcolor{gray}{76.2}$\pm$0.3 & 56.2$\pm$0.5 & \textcolor{gray}{58.9}$\pm$0.8 & 82.3$\pm$0.2 \\ 
        
        CR (hypo) & \textbf{84.8}$\pm$0.2 & {78.8}$\pm$0.5 & \textcolor{gray}{55.7}$\pm$0.4 & \textbf{59.4}$\pm$1.0 & 82.4$\pm$0.3\\
        
        PoE (lex) & \textcolor{gray}{82.9}$\pm$0.1 & \textcolor{gray}{74.3}$\pm$0.2 & \textcolor{gray}{55.8}$\pm$0.4 & \textcolor{gray}{58.4}$\pm$0.6 & \textcolor{gray}{81.7}$\pm$0.1\\
        
        PoE (hypo) & \textcolor{gray}{83.7}$\pm$0.4 & \textbf{79.1}$\pm$0.6 & \textcolor{gray}{55.6}$\pm$0.5 & 59.3$\pm$0.9 & \textcolor{gray}{81.4}$\pm$0.5\\
        
        BERT-aug & {84.7}$\pm$0.1 & {77.3}$\pm$0.2 & \textbf{56.5}$\pm$0.3 & \textbf{59.4}$\pm$0.6 & \textbf{82.6}$\pm$0.2 \\

        \bottomrule
		\end{tabular}
	\end{center}
	\caption{Comparing the impact of the augmentation to the confidence regularization (CR) \citep{utama2020mind}, and product-of-expert (POE) \citep{he-etal-2019-unlearn,clark-etal-2019-dont} methods debiased for the lexical overlap (lex) and hypothesis-only (hypo) biases. E.g., CR(lex) is the confidence regularization. All the results are reported on the mismatched subset of each evaluation set.}
	\label{tab:sota_comparison_mismatched}
\end{table*}

\section{Lexical Overlap Bias Model}
\citet{clark-etal-2019-dont} propose a simple model that uses a non-linear classifier on top of a set of simple features including: (1) whether all the hypothesis words exist
in the premise, (2) whether the hypothesis is a
subsequence of the premise, (3) the fraction
of hypothesis words that also exist in the premise, and (4) the
max and the average of the cosine distances between
the premise and hypothesis word vectors.
This model is used for detecting training examples that can be solved using the lexical overlap bias.

\section{Results on the mismatched evaluation sets}
Table~\ref{tab:sota_comparison_mismatched} presents the corresponding results of Table~\ref{tab:sota_comparison} on the mismatched subsets.

\end{document}